\RequirePackage{snapshot}
\documentclass[runningheads]{llncs}
\usepackage{graphicx}
\usepackage{wrapfig}
\begin{document}
%
\title{{PapagAI: \\ Automated Feedback for Reflective Essays}}
\titlerunning{PapagAI: \\ Automated Feedback for Reflective Essays}
\author{Veronika Solopova\inst{1}\orcidID{0000-0003-0183-9433} \and
Eiad Rostom\inst{1} \and Fritz Cremer\inst{1}\and
Adrian Gruszczynski\inst{1}\and Sascha Witte\inst{1}\and Chengming Zhang\inst{2}\orcidID{0009-0007-8695-5455}\and Fernando Ramos López\inst{1} \and Lea Plößl\inst{2}\orcidID{0009-0004-7290-5068}\and Florian Hofmann\inst{2}\and Ralf Romeike\inst{1}\orcidID{0000-0002-2941-4288}\and  Michaela Gläser-Zikuda\inst{2}\orcidID{0000-0002-3071-2995} \and Christoph Benzmüller\inst{2,3}\orcidID{0000-0002-3392-3093} \and Tim Landgraf\inst{1}\orcidID{0000-0003-4951-5235}}
\authorrunning{V. Solopova et al.}
%
\institute{Freie Universität Berlin, Germany \and
Friedrich-Alexander-Universität Erlangen-Nürnberg, Germany \and Otto-Friedrich-Universität Bamberg, Germany}
%
\maketitle              
%
%
\begin{abstract}

Written reflective practice is a regular exercise pre-service teachers perform during their higher education. 
Usually, their lecturers are expected to provide individual feedback, which can be a challenging task to perform on a regular basis. In this paper, we present the first open-source automated feedback tool based on didactic theory and implemented as a hybrid AI system. We describe the components and discuss the advantages and disadvantages of our system compared to the state-of-art generative large language models. The main objective of our work is to enable better learning outcomes for students and to complement the teaching activities of lecturers.

\end{abstract}
\section{Introduction}

Dropout rates as high as 83\% among pre-service teachers and associated teacher shortages are challenging the German education system \cite{dropoutrate,teachershortage}. This may be due to learning environments not adequately supporting prospective teachers in their learning process \cite{teacherdifficulties}. Written reflective practice may alleviate the problem: By reflecting on what has been learned and what could be done differently in the future, individuals can identify areas for improvement. However, instructors may be overburdened by giving feedback to 200+ students on a weekly basis. With the rise of large language models (LLMs, \cite{openai2023gpt4}), automated feedback may provide welcome relief. Students could iteratively improve their reflection based on the assessment of a specialized model and through that, their study performance. Instructors could supervise this process and invest the time saved in improving the curriculum. While current research is seeking solutions to align the responses of LLMs with a given set of rules, it is currently impossible to guarantee an output of a purely learnt model to be correct. Here, we propose ``PapagAI", a platform to write reflections and receive feedback from peers, instructors and a specialized chatbot. PapagAI uses a combination of ML and symbolic components, an approach known as hybrid AI \cite{Elands2019GoverningEA}. Our architecture is based on various natural language understanding modules\footnote{All ML models are available in our OSF depository (https://osf.io/ytesn/), while linguistic processing code can be shared upon request.}, which serve to create a text and user profile, according to which a rule-based reasoner chooses the appropriate instructions.


\section{Related work}
\label{rel}
PapagAI employs a number of models for detecting topics contained in -, and assessing the quality and depth of the reflection, as well as for detecting the sentiment and emotions of the author.  While extensive previous work was published on each of these tasks, implementations in German are rare. To our knowledge, there is no previous work that combined all in one application. Automated detection of reflective sentences and components in a didactic context has been described previously \cite{Geden2021PredictiveSM,Jung2020HowAH,pharmacy,Wulff2020ComputerBasedCO,ullman2019,10.1145/3170358.3170374}. In \cite{Jung2020HowAH}, e.g., the authors analyse the depth of a reflection on the text level according to a three-level scheme (none, shallow, deep). Document-level prediction, however, can only provide coarse-grained feedback. Liu et al.~\cite{LIU2021106733}, in contrast, also use three levels for predicting reflective depth for each sentence. 
In emotion detection, all previous works focus on a small set of 4 to 6 basic emotions. In Jena \cite{doi:10.1080/0144929X.2019.1625440}, e.g.,  the author describes detecting students' emotions in a collaborative learning environment. Batbaatar et al. \cite{8794541} describes an emotion model achieving an F1 score of 0.95 for the six basic emotions scheme proposed by Ekman~\cite{basic}. Chiorrini et al.~\cite{Chiorrini2021EmotionAS} use a pre-trained BERT to detect four basic emotions and their intensity from tweets, achieving an F1 score of 0.91. We did not find published work on the German language, except for Cevher et al.~\cite{cevher2019multimodal}, who focused on newspaper headlines. With regard to sentiment polarity, several annotated corpora were developed for German~\cite{potts,germevaltask2017}, mainly containing tweets. Guhr et al.~\cite{guhretal} use these corpora to fine-tune a BERT model. Shashkov et el. \cite{10.1145/3430895.3460150} employ sentiment analysis and topic modelling to relate student sentiment to particular topics in English. Identifying topics in reflective student writing is studied by Chen et al.~\cite{10.1145/2883851.2883951} using the MALLET toolkit~\cite{McCallum2022} and by De Lin et al.~\cite{9637181} with Word2Vec + K-Means clustering. The techniques in these studies are less robust than the current state-of-art, such as ParlBERT-Topic-German \cite{klamm2022frameast} and Bertopic~\cite{Grootendorst2022BERTopicNT}. Overall, published work on automated feedback to student reflections is scarce, the closest and most accomplished work being AcaWriter \cite{KnightVijay} and works by Liu and Shum \cite{LIU2021106733}. They use linguistic techniques to identify sentences that communicate a specific rhetorical function. They also implement a 5-level reflection depth scheme and extract parts of text describing the context, challenge and change. The feedback guides the students to the next level of reflective depth with a limited number of questions. In their user study, 85.7\% of students perceived the tool positively. However, the impact on the reflection quality over time was not measured and remains unclear.

\section{Methods, components and performances}
\label{met}
\paragraph{Data collection.} Our data comes from the German Reflective Corpus \cite{solopova-etal-2021-german}. The dataset contains reflective essays collected via google-forms from computer science and ethics of AI students in German, as well as e-portfolio diaries describing school placements of teacher trainees from Dundee University.
 For such tasks as reflective level identification and topic modelling, we enlarged it by computer science education students' essays and pedagogy students' reflections\footnote{This still non-published data can be obtained upon request.}. It consists of reflections written by computer science, computer science education, didactics and ethics of AI students in German and English. Data is highly varied, as didactics students write longer and deeper reflections than e.g.~their computer science peers. 

\vspace*{-.2em}

\vspace*{-.2em}
\paragraph{Emotions detection.} Setting out from the Plutchik wheel of basic emotions \cite{plutchik}, during the annotation process we realised that many of the basic emotions are never used, while other states are relevant to our data and the educational context (e.g. confidence, motivation). We framed it as a multi-label classification problem at the sentence level. We annotated 6543 sentences with 4 annotators.
The final number of labels is 17 emotions, with the 18th label being 'no-emotion'.
We calculated the loss using binary cross entropy, where each label is treated as a binary classification problem, the loss is calculated for each label independently, which we sum for the total loss. We achieved the best results with a pre-trained RoBERTa \cite{Liu2019RoBERTaAR}~, with a micro F1 of 0.70 and a hamming score of 0.67 across all emotion labels. The model achieved the highest scores for ``surprise”, ``approval” and ``interest”. With a lenient hamming score, accounting for the model choosing similar emotions (e.g. disappointment instead of disapproval) our model achieves up to 0.73.

\vspace*{-.2em}
\paragraph{Gibbs cycle.} \cite{Gibbs1988-hl} illustrates cognitive stages needed for optimal reflective results. It includes 6 phases: \textit{description}, \textit{feelings}, \textit{evaluation}, \textit{analysis}, \textit{conclusion} and \textit{future plans}. We annotated the highest phase present in a sentence and all the phases present.
We treated this as a multi-class classification problem and used a pre-trained ELECTRA model. While evaluating, we compared one-hot prediction to the highest phase present and 3 top probability classes with all the phases present. While one-hot matching only managed to score 65\% F1 macro, the top 3 predictions achieve up to 98\% F1 macro and micro.

\vspace*{-.2em}
\paragraph{Reflective level detection.} Under the supervision of Didactics specialists two annotators labelled 600 texts according to Fleck \& Fitzpatrick's scheme \cite{fleck}, achieving moderate inter-annotators agreement of 0.68. The coding scheme includes 5 levels: \textit{description}, \textit{reflective description}, \textit{dialogical reflection}, \textit{transformative reflection} and \textit{critical reflection}; With 70\% of the data used for the training and 30\% for evaluation, we used pre-trained BERT large and complete document embeddings for the English and German, resulting in QWK score of 0.71 in cross-validation.
\vspace*{-.2em}
\paragraph{Topic modelling.} We used BERTopic \cite{Grootendorst2022BERTopicNT} on the sentence level. First, we tokenized and normalize the input sequence to lowercase and filter out numbers, punctuation, and stop-words using nltk library \cite{bird2009natural}. Then, we extract embeddings with BERT, reduce dimensionalities with UMAP, cluster reduced embeddings with HDBSCAN, create topic representation with tfidf and fine-tune topic representations with the BERT model. Because we have a lot of data of different origins, we created two clusterings, one more specific to the pedagogy topic and one including various educational topics. You can see our clusters in App.
\vspace*{-.2em}
\paragraph{Linguistic scoring.} Using spacy\footnote{https://spacy.io} 
we tokenized, and lemmatize the sentences, extracted dependencies parcing and part of speech. Additionally, we used RFTagger\cite{Schmid2008-hy} for parts of speech and types of verbs. We extract sentence length, adverb for verb ratio, adjective for noun ratio, number of simple and complex sentences, types of subordinated clauses and number of discourse connectors\footnote{We use Connective-Lex list for German:  {https://doi.org/10.4000/discours.10098}.} used. This information enables us to determine the reflection length, expressivity and variability of the language, as well as surface coherence and structure.

\section{System architecture}
\label{arch}
\begin{figure}[tp]
\centering 
\includegraphics[width=8.3cm]{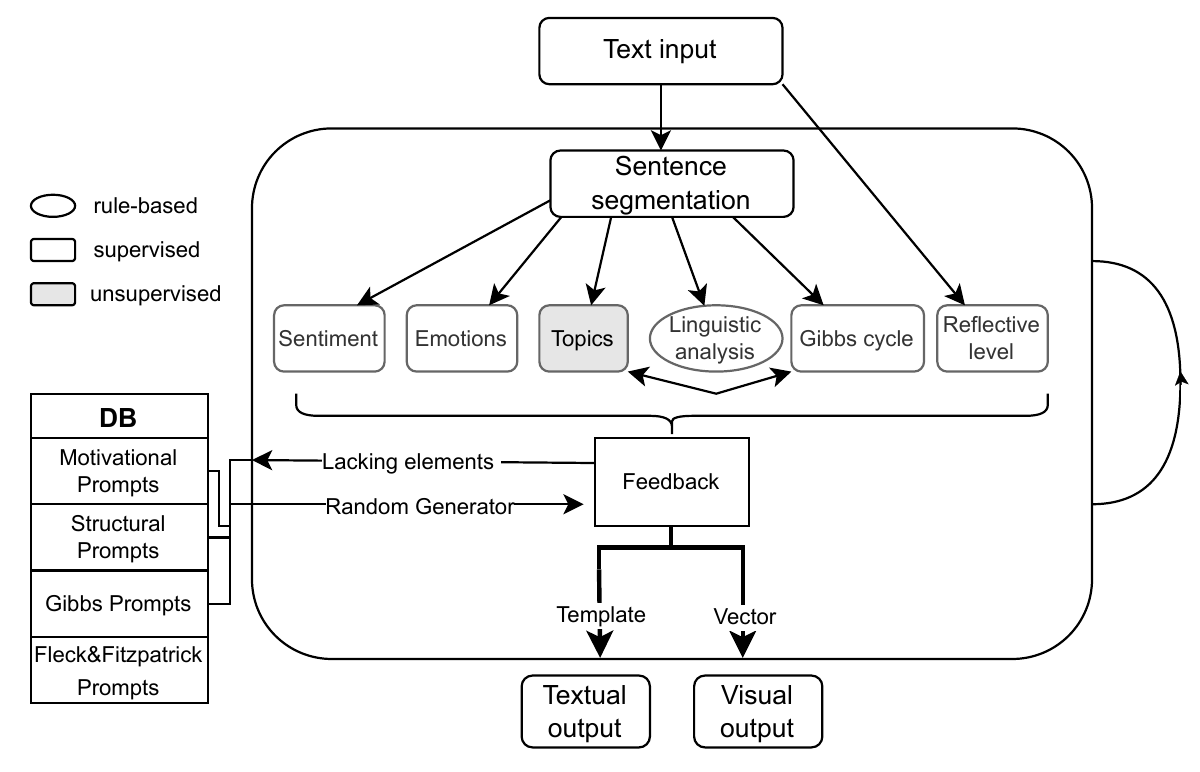}
\caption{The diagram of our PapagAI system shows the main productive modules. The legend on the left indicates the nature of the AI modules used.} \label{fig1}
\end{figure}
\begin{figure}[tp]
\centering 
\includegraphics[width=8.3cm]{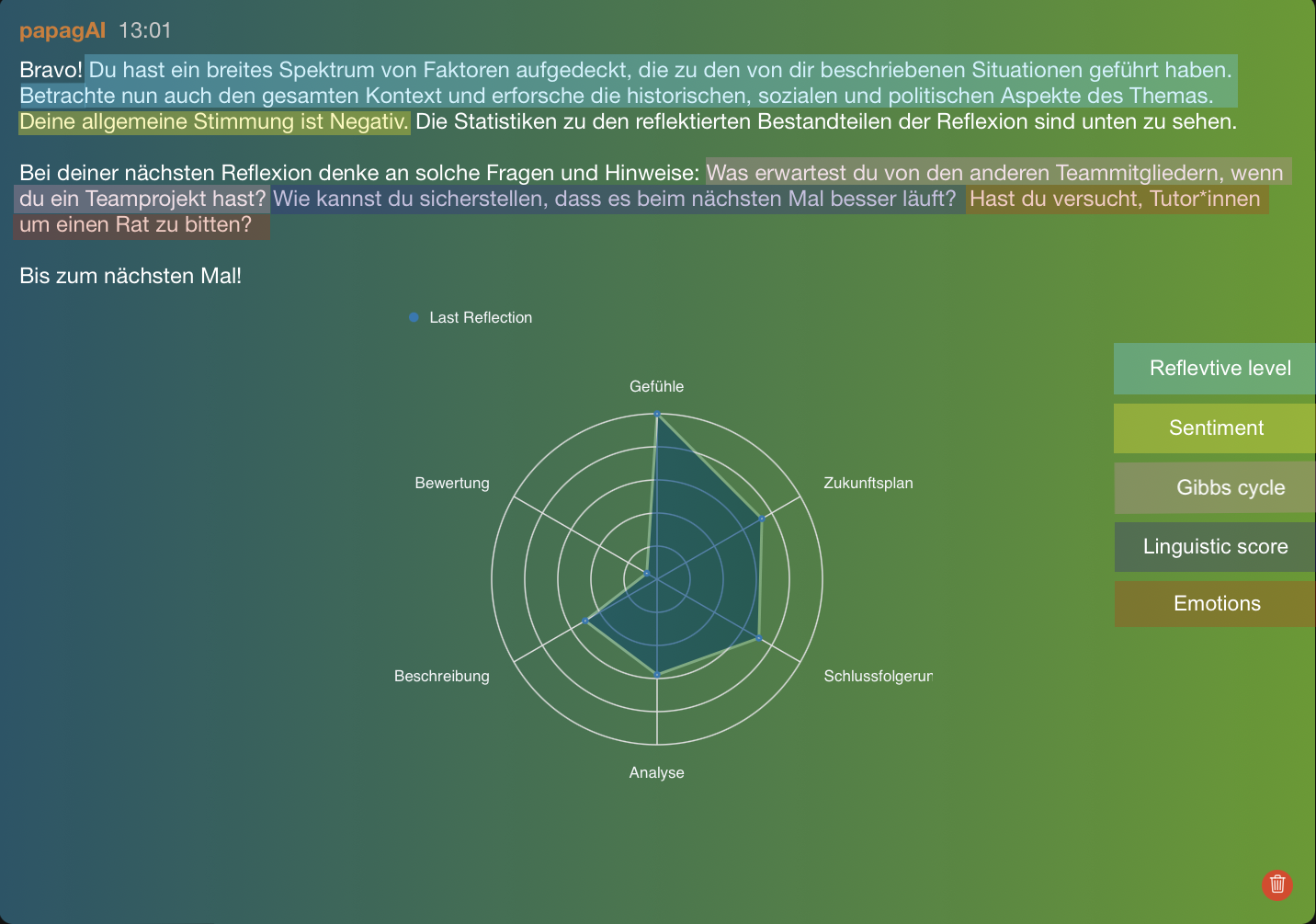}
\caption{The radar below the textual feedback illustrates Gibbs cycle completeness. The colour of the highlighted feedback text corresponds to the model responsible for this information.} \label{fig2}
\end{figure}

In PapagAI (see Fig.~\ref{fig1}) 
the input text of the reflection is received from the AWS server through a WebSocket listener script. To minimize the response time, the models are loaded in the listener script once and then the user request spawn threads with the models already loaded. If the input text is smaller than three sentences and contains forbidden sequences, the processing does not start and the user receives a request to revise their input. Otherwise, the text is segmented into sentences and tokens. The language is identified using langid \cite{lui-baldwin-2012-langid} and if the text is not in German, it is translated using Google translator API implementation.\footnote{https://pypi.org/project/deep-translator/} The reflective level model receives the whole text, while other models are fed with the segmented sentences. Topic modelling and Gibbs cycle results are mapped, to identify if topics were well reflected upon. If more than three sentences are allocated to the topic and these sentences were identified by the Gibbs cycle model as analysis, we consider the topics well thought through. The extracted features are then passed to the feedback module. Here, the lacking and under-represented elements are identified in linguistic features and the three least present Gibbs cycle stages. If sentiment and emotions are all positive we conclude that no potential challenges and problems are thought through. If the sentiment and emotions are all negative, we want to induce optimism. These features together with the reflective level are mapped to the database of potential prompts and questions, where one of the suitable feedback options is chosen randomly for the sake of variability. Using manually corrected Gpt-3 outputs, for each prompt we created variations so that the feedback does not repeat often even if the same prompts are required.The extracted textual prompts are built together in a rule-based way into the template, prepared for German, Spanish and English. Otherwise, the overall feedback is made in German and then translated into the input language. The textual and a vector of extracted features for visual representation are sent back to the AWS server. The whole processing takes from 15 to 30 seconds based on the length of the text. Sample feedback can be seen in Figure \ref{fig2}.
\vspace*{-.2em}
\section{Comparison with GPT-3}
\label{gpt}
\vspace*{-.2em}
We compared our emotions detection (fine-tuned RoBERTa) and Gibbs cycle model (fine-tuned Electra) with the prompt-engineered state-of-the-art generative model Davinci \cite{brown2020language} on the same task. For the evaluation and comparison, we used a small subset of 262
samples which were not part of the training. 
We first tried the zero-shot approach, where we described our labels to GPT-3 and gave it our sentence to predict. Then, we tried a one-shot approach, providing GPT-3 with one example sentence for each label. Finally,  in the few-shot approach, we provided GPT-3 with three examples per label, which is the maximum number of examples possible due to the input sequence length restriction. Although the task requested GPT-3 to pick multiple labels out of the possible options, the model predicted multiple labels only in 5\% of the cases for emotions. For this reason, we used the custom defined ``one correct label”: the score considers the prediction correct if it contains at least one correct label from the sentence's true labels.
The zero-shot approach achieved only 0.28 accuracy in predicting one correct label for emotions. The model predicted the labels ``information”, ``uncertainty”, ``interest”, and ``motivated” for the
majority of the sentences. With the Gibbs cycle task, it achieved 80\% correct predictions. Providing one example per label improved the performance noticeably by 18\% (0.46) for emotions, and
the model was able to detect emotions like ``confidence”, ``challenged”, and
``approval” more accurately. It did not influence Gibb's cycle performance. Increasing the number of examples to three resulted in a slight improvement of 3\% (0.49) for emotions, and 7\% (0.87) for the Gibbs cycle. However, the best-scoring approaches did not offer a comparable performance to our fine-tuned models on these specific tasks with 0.81 on the same custom metric for emotion detection and 0.98 for the Gibbs cycle.

\section{Discussion and conclusion}\label{disc}

 
The current PapagAI system has several advantages in comparison to generative LLMs. It ensures transparency of the evaluation and control over the output, which is based exclusively on didactic theory. Although LLMs show huge promise, they are still prone to hallucination \cite{Ji2023,manakul2023selfcheckgpt}, and, as we have shown in \S\ref{gpt}, they may under-perform on difficult cognitive tasks in comparison to smaller language models fine-tuned for the task. The fine-tuning of LLMs to didactic books and instructions, which we plan for our future work, still does not guarantee 100\% theoretical soundness of the output, which is problematic e.g.~in the case of pre-service students with statistically low AI acceptance. At the same time, the newest models, such as GPT-4, are only available through APIs, which raises concerns about data privacy, especially as the data in focus is an intimate reflective diary. Moreover, current open-source models, such as GPT-J and GPT-2, especially for languages other than English do not draw comparable results. Our architecture has, however, obvious drawbacks. On the one hand, our models do not reach 100\% accuracy and this can naturally lead to suboptimal feedback. The processing time for many models, especially for longer texts, can be significantly higher than for a single generative LLM. For now, as we provide one feedback message for one rather long reflection, this is not a big issue, however, if we implement a dialogue form, the time of response would not feel natural. Finally, the variability of output using our approach is much more limited in comparison to generative models. We try to address it by creating many similar versions of instructions rephrased by GPT-3, and corrected manually. On average 7 out of 10 prompts needed some correction. Most of the errors were related to GPT-3 trying to rephrase the given sentence using synonyms that were not didactically appropriate in the given context. 
Future work, among others, will focus on user studies to understand how we can optimize the feedback, so that the users find it credible and useful, while their reflective skills advance. We also plan a more detailed evaluation based on more user data. We hope that our work will contribute to the optimization of the pre-service teachers' reflective practice and self-guided learning experience.

\bibliographystyle{splncs04}
\bibliography{papagai}

%
%
%
\clearpage\section*{Appendixes}

\begin{table}
\caption{Metrics mentioned in the paper.}\label{tab1}
\begin{tabular}{|l|l|}
\hline
Metric & Definition\\
\hline
 F1-score &  A harmonic mean of the precision and recall calculated per class. Can range \\&  from 0 to 1. \\\hline
 F1-score macro & The metric is computed independently for each class and then the average \\& is taken. \\\hline
 F1-score micro &  The metric aggregates the contributions of all classes to compute the average \\&  metric. \\\hline
Cohen's kappa &  The metric is used to measure inter-annotator reliability for categorical items. \\&  0.41–0.60 is 
 interpreted as moderate agreement, 0.61–0.80 as substantial, and  \\& 0.81–1.00 as perfect agreement. \\\hline
QWK &  Quadratic Weighted Kappa measures the agreement between two outcomes \\&  ranging from -1  (complete disagreement) to 1 (complete agreement).\\\hline
Hamming score &  The metric is often used for multi-label classification calculating the fraction \\& of wrong labels to the total number of labels. The values higher than 0.9  \\& are excellent scores, higher than 0.7 are good scores, and lower than 0.7 may \\& be considered poor.\\
\hline
\end{tabular}
\end{table}

\begin{table}
\caption{Topics clusters from Bertopic.}\label{tab2}
\begin{tabular}{|l|l|}
\hline
Clustering 1& Clustering 2\\
\hline
 Lectures and editing& Teamwork and Tasks\\
Classroom Management&Teacher, school, teaching\\
Pedagogy and Educational Diagnostics&Algorithms, Computer Science, Digital Technology\\
Reading and Literature & Self-promotion\\
Conflict Analysis & Music\\
Feedback & Math and numeracy\\
Your Subject Area & Science and Experiments\\
Diagnostics and diagnostic procedures & \\
Intervention measure &\\
Motivation &\\
Portfolio &\\
Lecture material and video &\\
Psychology &\\
\hline
\end{tabular}
\end{table}

\begin{table}
\centering
\caption{Emotion detection labels.}\label{tab3}
\begin{tabular}{|l|}
\hline
Emotions \& Feelings \\
\hline
information\\
annoyance\\
 appreciation\\
disapproval/critique\\
interest \\
anticipation\\
excitement\\
challenged\\
confidence\\
disappointment\\
insecurity\\
motivation\\
optimism\\
responsibility\\
satisfaction\\
surprise\\
uncertainty\\
wariness\\
\hline
\end{tabular}
\end{table}

\begin{table}
\caption{Defenitions of reflective labels.}\label{tab4}
\begin{tabular}{|l|l|}
\hline
Level &  Defenition\\
\hline
Description & It is the lowest level, where the person only describes the \\
& circumstances and may include an evaluation of their own feelings.\\
\hline
Reflective description & Here one's own perspective analysis and superficial justifications \\
& are present.\\
\hline
Dialogical Reflection & It includes analysis of various perspectives as if in form of \\& an internal dialogue with oneself.\\
\hline
Transformative Reflection & It should include the plan for the next steps or \\
& what one would do next time in such a situation. \\
\hline
Critical Reflection & The highest level of reflection encompasses a wider\\ 
& context (social, political, historical).\\
\hline
\end{tabular}
\end{table}






\end{document}